\documentclass[10pt,a4papert]{article}
\usepackage[utf8]{inputenc}
\usepackage[english]{babel}
\usepackage{authblk}
\usepackage{hyperref}
\usepackage{graphicx}
\usepackage{amsmath}
\usepackage{booktabs}
\usepackage{listings}
\usepackage{xcolor}
\usepackage{caption}
\usepackage{subcaption}
\usepackage{tabularx}
\usepackage{siunitx}
\usepackage{tcolorbox}
\usepackage{url}
\usepackage[numbers,sort&compress]{natbib}

\hypersetup{
    colorlinks=true,
    linkcolor=blue,
    filecolor=magenta,      
    urlcolor=cyan,
    pdftitle={6ABOS: EnMAP Atmospheric Correction},
}

\title{\textbf{6ABOS: An Open-Source Atmospheric Correction Framework for the EnMAP Hyperspectral Mission Based on 6S}}

\author{
    Gabriel Caballero Cañas$^{1,*}$, 
    B\'{a}rbara Alvado Arranz$^{1}$, 
    Xavier S\`{o}ria-Perpiny\`{a}$^{1}$,
    Antonio Ruiz-Verd\'{u}$^{1}$, 
    Jes\'{u}s Delegido$^{1}$, 
    and Jos\'{e} Moreno$^{1}$
    \vspace{0.3cm} \\ 
    \small $^{1}$Laboratory for Earth Observation (LEO), Image Processing Laboratory (IPL), \\
    \small Universitat de Val\`{e}ncia, Spain \\
    \small $^{*}$Corresponding author: 
    gabriel.caballero@uv.es
}
\date{\today}

\begin{document}

\maketitle

\begin{abstract}
The Environmental Mapping and Analysis Program (EnMAP) mission has opened new frontiers in the monitoring of optically complex environments. However, the accurate retrieval of surface reflectance over water bodies remains a significant challenge, as the water-leaving signal typically accounts for only a small fraction of the total radiance, being easily obscured by atmospheric scattering and surface reflection effects. This paper introduces \textit{6ABOS} (6S-based Atmospheric Background Offset Subtraction), a novel open-source Python framework designed to automate the atmospheric correction (AC) of EnMAP hyperspectral imagery. By leveraging the Second Simulation of the Satellite Signal in the Solar Spectrum (6S) radiative transfer model, \textit{6ABOS} implements a physically-based inversion scheme that accounts for Rayleigh scattering, aerosol interactions, and gaseous absorption. The framework integrates automated EnMAP metadata parsing with dynamic atmospheric parameter retrieval via the Google Earth Engine (GEE) Application Programming Interface (API). Validation was conducted over two Mediterranean inland water reservoirs with contrasting trophic states: the oligotrophic Benagéber and the hypertrophic Bellús. Results demonstrate a high degree of spectral similarity between in situ measurements and EnMAP-derived water-leaving reflectances. The Spectral Angle Mapper (SAM) values remained consistently low ($SAM < 10^\circ$) across both study sites.%

~\textit{6ABOS} is distributed via \textit{conda-forge}, providing the scientific community with a scalable, transparent, and reproducible open-science tool for advancing hyperspectral aquatic research in the cloud-computing era.
\end{abstract}

\vspace{0.5cm}

\section{Introduction}
\subsection{The Atmospheric Correction Challenge in Hyperspectral Aquatic Remote Sensing}

Inland water bodies are vital ecosystems for biodiversity and human health~\cite{biggs2017}; however, monitoring these environments via remote sensing presents significant scientific challenges~\cite{karpatne2016}. Principally, aquatic remote sensing is constrained by a low signal-to-noise ratio ($SNR$): water-leaving radiance is often masked by atmospheric path radiance, consisting of both the diffuse radiance scattered by atmospheric constituents (aerosols and gas molecules) and the environmental contribution, which occurs when radiance reflected from the target’s surroundings is subsequently scattered by the atmosphere into the sensor’s field of view~\cite{Nair1997}. This atmospheric path radiance can account for up to 90\% of the signal reaching the Top-of-Atmosphere (TOA). Consequently, the resulting signal captured by spaceborne sensors exhibits low contrast, as the relevant information concerning water's Optically Active Components (OACs) is often restricted to less than 20\% of the total detected radiance~\cite{hovis1977}. Furthermore, the retrieval of OACs in complex waters requires a delicate balance between spatial and spectral resolution—a trade-off that traditional multispectral sensors have historically struggled to overcome~\cite{olmanson2011}. 

The~\textbf{Environmental Mapping and Analysis Program (EnMAP)}~\cite{Stuffler2007} addresses this limitation. EnMAP features a push-broom imaging spectrometer covering the 420--2450~nm range with 228 spectral channels. Its high $SNR$ ($> \text{400:1}$) in the Visible and Near-Infrared (VNIR) region~\cite{Guanter2015} and 30~m spatial resolution facilitate the discrimination of fine-scale spectral features, which is paramount for disentangling the overlapping signals of OACs in optically complex inland waters.

However, the high spectral fidelity of EnMAP can only be fully exploited if the atmospheric contribution is accurately decoupled from the surface signal. Despite the sensor's advanced capabilities, the transition from TOA radiance to water-leaving reflectance ($\rho_w$) remains the most critical step in the processing chain. This dimensionless parameter, defined as the ratio of water-leaving radiance to total downward irradiance, is ideally normalised to account for a hypothetical overhead Sun and the absence of atmospheric attenuation~\cite{Gordon1980}.

In aquatic scenes, the atmospheric signal is not merely a noise factor but a dominant component that varies dynamically in space and time. Consequently, the process of \textit{Atmospheric Correction} (AC) is critical in this context. In addition, in inland waters, this task is further complicated by:

\begin{itemize}
    \item \textbf{Adjacency Effects:} Signal contamination from surrounding terrestrial vegetation and soil, often referred to as environmental radiance.
    \item \textbf{Optical Complexity:} high Coloured Dissolved Organic Matter (CDOM) absorption in the blue-green spectrum and non-zero Near-Infrared (NIR) reflectance in turbid waters, which invalidates the "dark pixel" assumption for AC. If one assumes that the NIR signal is caused solely by atmospheric contributions and extrapolates these measurements to correct visible bands, essential components of the signal are removed~\cite{moore1999}.
    \item \textbf{Surface Effects:} Specular reflections, including sun and sky glint, that distort the water-leaving signal.
\end{itemize}

\subsection{State of the Art: Atmospheric Correction for EnMAP in Aquatic Studies}

The scientific community currently relies on several processors to derive $\rho_w$ from EnMAP data. In recent years, several AC processors have been adapted or specifically developed to handle the high dimensionality of EnMAP data in aquatic environments. The following section provides a summary of the current state-of-the-art AC processors available for EnMAP imagery in aquatic research.

The~\textbf{EnMAP L2A}~\cite{Kiselev2015} processor utilises the \textit{Modular Inversion Program} (MIP), trained with MODTRAN-5 and a multicomponent radiative transfer model specifically for aquatic environments. Unlike standard land processors, it adjusts the Aerosol Optical Depth (AOD), the Water Vapour (WV), and water components simultaneously to find the best spectral match. It includes specific modules for sun glint, haze/cirrus detection, and adjacency correction.

~\textbf{ACOLITE} (Atmospheric Correction for OLI 'lite')~\cite{Vanhellemont18} employs the \textit{Dark Spectrum Fitting} (DSF) algorithm~\cite{Vanhellemont20}. It assumes that water reflectance in the Short-Wave Infrared (SWIR) region is negligible, attributing the signal in these bands to atmospheric path radiance. ACOLITE is particularly robust in high turbidity conditions and iteratively estimates WV from initial values.

~\textbf{POLYMER} (POLYnomial-based algorithm applied to MERiS)~\cite{steinmetz2011atmospheric} has been successfully adapted for hyperspectral sensors, utilising a spectral decomposition method that effectively handles sun glint and maritime aerosols.

~\textbf{PACO} (Python-based Atmospheric Correction) processor~\cite{paco2020}, developed by the German Aerospace Agency (DLR), is based on the MODTRAN radiative transfer code and is specifically engineered to handle the complex geometries of hyperspectral push-broom sensors. When applied to aquatic scenes, PACO is often coupled with the WASI (Water Colour Simulator)~\cite{wasi2004} module. This combination allows for a simultaneous inversion of atmospheric and water constituents.

Table~\ref{tab:ac_comparison} presents a comparative overview of the AC frameworks currently applicable to EnMAP imagery.

\begin{table}[htbp]
    \centering
    \caption{Summary of prominent AC processors currently utilised for EnMAP aquatic research.}
    \label{tab:ac_comparison}
    \small
    \begin{tabularx}{\textwidth}{l p{3.5cm} X p{2.5cm}}
        \toprule
        \textbf{Processor} & \textbf{Algorithm Type} & \textbf{Primary Advantage} & \textbf{Aquatic Context} \\ 
        \midrule
        \textbf{EnMAP L2A} & Physical Inversion (MIP) & Operational standard; integrated workflow. & Optimized for land; limited in water. \\
        \addlinespace
        \textbf{ACOLITE} & DSF & Robust aerosol retrieval without NIR-zero assumption. & Excellent for turbid inland waters. \\
        \addlinespace
        \textbf{POLYMER} & Spectral Decomposition & Effective sun glint and maritime aerosol removal. & High performance in coastal/oceanic. \\
        \addlinespace
        \addlinespace
        \textbf{PACO/WASI} & MODTRAN / Inversion & DLR native; joint atmosphere-water inversion. & Operational EnMAP. \\
        \bottomrule
    \end{tabularx}
\end{table}

\subsection{Research Rationale and Objectives}

The current landscape of aquatic remote sensing is undergoing a paradigm shift with the advent of the EnMAP hyperspectral mission. While this sensor provides the spectral density required to resolve complex water constituents, achieving reliable AC outcomes continues to be a persistent challenge that warrants extensive community-driven research and methodological innovation. To bridge this gap, this study introduces \textbf{\textit{6ABOS}} (6S-based Atmospheric Background Offset Subtraction)~\cite{caballero_canas_2026_18300277}, an emergent open-source Python framework that automates AC by leveraging the Second Simulation of the Satellite Signal in the Solar Spectrum (6S)~\cite{vermote1997atmospheric, vermote1997second} Radiative Transfer Model (RTM) to retrieve $\rho_{w}$ from EnMAP imagery over inland water bodies. \textit{6ABOS} provides a modular tool that enables physical simulations of atmospheric effects\,---\,including Rayleigh scattering, water vapour absorption, and aerosol interactions\,---\,which must be decoupled from the TOA signal to isolate $\rho_{w}$. By combining the proven accuracy of the 6S model with a transparent, modular architecture, \textit{6ABOS} offers a robust alternative for the scientific community to advance hyperspectral aquatic research.

Beyond the introduction of~\textit{6ABOS}, this study specifically aims to:

\begin{enumerate}
    \item~Develop a modular architecture that integrates the 6S RTM with the automated retrieval of dynamic atmospheric auxiliary data.
    
    \item~Implement an optimized processing scheme capable of handling hyperspectral data cubes efficiently.
    
    \item~Validate the spectral performance of~\textit{6ABOS} across oligotrophic and hypertrophic Mediterranean inland waters by comparing retrieved $\rho_{w}$ against in situ radiometry.
    
    \item~Provide an open-science, Python-based tool that ensures reproducibility and facilitates integration for the scientific community engaged in hyperspectral aquatic monitoring.
\end{enumerate}

\section{Physical Model and Methodology}
\subsection{The 6S Radiative Transfer Model}

For the atmospheric correction phase, the \textit{6SV (Version 1.1)} radiative transfer code was selected by its ability to perform vectorial radiation treatment, enabling consistent incorporation of polarisation via the Stokes parameters. This feature provides \textit{6SV} with a significant physical advantage over scalar models, such as MODTRAN~\cite{MODTRAN}, as polarisation is a critical factor for accurate AC.

While the 6S code is inherently formulated as a forward radiative transfer model (from surface to TOA), previous studies have reported that inversion strategies based solely on pre-computed forward-mode lookup tables (LUTs) may introduce non-negligible interpolation errors, particularly under high-gradient atmospheric conditions and strongly nonlinear parameter regimes. Such limitations arise from the finite sampling of high-dimensional atmospheric state spaces, where conventional LUT interpolation may fail to adequately represent rapid variations in radiative transfer behaviour~\citep{Cremer2025,verrelst2025rtm,lamminpaa2025forward}.

In contrast, the proposed \textit{6ABOS} framework does not rely on static forward LUT inversion. Instead, it employs 6S as a physically-based radiative transfer engine to dynamically simulate the atmospheric state 
for each acquisition geometry and spectral channel. Within this scheme, 6S is used to explicitly derive the key atmospheric parameters governing the radiative transfer equation, including direct and diffuse solar irradiances, atmospheric path radiance, gaseous transmittance, upward target-to-sensor total transmittance, and the spherical albedo of the atmosphere.

By parameterising the 6S core on a per-scene and per-band basis, 
\textit{6ABOS} performs an explicit inversion of the radiative transfer 
equation, thereby converting TOA radiance into Bottom-of-Atmosphere (BOA) reflectance through a physically consistent 
solution. This strategy mitigates the interpolation artefacts reported in purely forward-mode LUT approaches.

A critical aspect of this integration is the spectral discretisation. While the 6S model allows for user-defined sampling, it operates at a fundamental spectral resolution limit of $\Delta\lambda = 2.5$ nm. Given that the EnMAP Spectral Response Functions (SRFs) exhibit a Full Width at Half Maximum (FWHM) of approximately $6.5$ nm in the VNIR region, our processing grid satisfies the Nyquist-Shannon sampling criterion (see Equation~\ref{eq:nyquist}), as the sampling step is finer than half the instrument's bandwidth:

\begin{equation}
    \Delta\lambda_{6S} \le \frac{FWHM_{EnMAP}}{2} \approx 3.25 \text{ nm}
\label{eq:nyquist}
\end{equation}

Although the spectral sampling strategy satisfies the Nyquist–Shannon criterion for the EnMAP SRFs, it is important to acknowledge several structural simplifications inherent in the 6S RTM~\cite{Cremer2025}.

First, the intrinsic spectral resolution of 6S remains relatively limited. While this sampling is generally sufficient for broadband atmospheric simulations, it may not fully resolve narrow
absorption structures, particularly within strong gaseous absorption bands. Consequently, the effective spectral representation results from the combined interaction between the model sampling grid and the sensor SRFs, which may introduce minor discrepancies in spectral sampling and
band-shape representation.

Second, the atmospheric vertical structure in 6S is represented through a layered profile in which different atmospheric constituents are treated in
separate blocks (e.g., first aerosols, then water vapour, then ozone). Such an ordered
stratification does not fully reflect the continuous mixing observed in real atmospheric columns. In practice, the ordering of these layers may slightly affect the simulated radiative transfer solution, highlighting a structural
simplification of the model formulation.

Third, the radiative transfer scheme in 6S treats scattering and absorption processes through an explicit separation. While this approach enables
computational efficiency, it becomes less accurate within strong gaseous absorption bands.

Despite these limitations, the relative simplicity and numerical stability of 6S remain advantageous for operational AC workflows.
In particular, its physically grounded formulation, low computational cost,
and long-standing validation across multiple satellite missions make it a
practical and reliable RTM for routine processing of hyperspectral imagery.
Within the \textit{6ABOS} framework, these characteristics allow efficient
per-band atmospheric simulations while maintaining a transparent and reproducible processing chain suitable for large-scale aquatic monitoring
applications.

While the current model offers a robust physical description of scattering processes, a future development would be to integrate absorption files with higher spectral resolution. This enhancement would extend the benefits of the \textit{6SV} vectorial framework to hyperspectral applications, optimising the modelling of gaseous absorption bands without compromising the fundamental integrity of the original code.

The workflow implemented in this framework follows a physical inversion scheme:

\begin{enumerate}
    \item \textbf{$O_{3}$ Correction:} The measured $L_{TOA}$ is initially corrected for $O_{3}$ absorption by applying the gaseous transmittance factor $T_{g,O_{3}}$.
    \item \textbf{Path Radiance Subtraction:} The atmospheric path radiance ($L_{path}$), which represents photons scattered directly by the atmosphere toward the sensor, is modelled with 6S and subtracted from the signal reaching the sensor at TOA.
    \item \textbf{Normalisation and Transmittance:} The $O_{3}$-corrected signal is normalised by the total downwelling solar irradiance ($E_{s}$) and the upward (surface-to-sensor) transmittance ($T_{\uparrow}$), which accounts for both direct and diffuse components.
    \item \textbf{Atmospheric Coupling:} The formulation accounts for the reflections between the surface and the atmosphere via the spherical albedo ($S_{atm}$), decoupling the target reflectance from the diffuse atmospheric background. 
    \item \textbf{Gaseous Transmittance Masking:} The total gaseous transmittance ($T_{g}$) is evaluated for each spectral channel. If the transmittance falls below a user-defined threshold, the channel is flagged and excluded from the final data product. This prevents noise amplification in spectral regions characterised by low atmospheric transmissivity, such as deep water vapour or oxygen absorption bands.
\end{enumerate}

The fundamental radiative transfer equation for the BOA $\rho_{w}$ is:

\begin{equation}
    \rho_{w}(\lambda) = \frac{\left( \frac{L_{TOA}(\lambda)\,d^2}{T_{g,O_{3}}(\lambda)} - L_{path}(\lambda) \right)}{\frac{E_{s}(\lambda) \cdot T_{\uparrow}(\lambda)}{\pi} + S_{atm}(\lambda) \left( \frac{L_{TOA}(\lambda)\,d^2}{T_{g,O_{3}}(\lambda)} - L_{path}(\lambda) \right)}
\label{eq:inversion}
\end{equation}

Where $L_{\text{TOA}}(\lambda)$ represents the radiance at TOA; $d$ is the Earth-Sun distance in astronomical units (AU); $T_{g,\text{O}_{3}}(\lambda)$ is the ozone transmittance factor; $L_{\text{path}}(\lambda)$ denotes the atmospheric path radiance, accounting for both Rayleigh and aerosol scattering; $E_{s}(\lambda)$ is the solar irradiance (direct + diffuse), $T_{\uparrow}(\lambda)$ refers to the upward atmospheric transmittance; and $S_{\text{atm}}(\lambda)$ is the spherical albedo of the atmosphere. All the aforementioned variables are spectral magnitudes fundamentally dependent on the wavelength ($\lambda$), as denoted by the spectral notation.
In aquatic remote sensing, the primary goal of AC is to retrieve the water-leaving signal, which carries the information about the water's OACs. To ensure comparability across different illumination conditions and atmospheric states, this signal is typically expressed as Remote Sensing Reflectance ($R_{rs}$)~\cite{mobley1994light}. The $R_{rs}$ ($sr^{-1}$) is defined as the ratio of the water-leaving radiance to the total downwelling irradiance reaching the water surface (Equation~\ref{eq:rrs},~\cite{Mobley99}):

\begin{equation}
R_{rs}(\lambda) = \frac{L_{w}(\lambda, 0^{+})}{E_{d}(\lambda, 0^{+})} \approx \frac{\rho_{w}(\lambda)}{\pi}
\label{eq:rrs}
\end{equation}

Where $L_w(\lambda)$ is the water-leaving spectral radiance ($W \, m^{-2} \, sr^{-1} \, nm^{-1}$); $E_d(\lambda)$ is the downwelling spectral plane irradiance incident onto the water surface ($W \, m^{-2} \, nm^{-1}$); $\rho_w(\lambda)$ is the dimensionless spectral water-leaving reflectance; and $0^+$ indicates that the quantities are derived just above the air-water interface.

This formulation assumes a Lambertian approximation for the water-leaving radiation field, where the signal is considered isotropic across the upward hemisphere. Consequently, the $\rho_{w}$ is directly related to $R_{rs}$ by the factor $\pi$, such that $R_{rs}(\lambda) = \rho_w(\lambda) / \pi$, which allows for a standardised representation of the aquatic optical properties.

\subsection{Parameterization for EnMAP}
\textit{6ABOS} performs an automated parsing of the EnMAP XML metadata to parameterize the radiative transfer core. This ensures a scene-average characterisation of the following variables:

\begin{itemize}
    \item \textbf{Acquisition Geometry and Date:} Precise extraction of the Sun Zenith Angle (SZA), Sun Azimuth Angle (SAA), View Zenith Angle (VZA), and View Azimuth Angle (VAA) from the metadata. Additionally, the scene acquisition date is retrieved to calculate the Julian Day, which is used to determine the Earth-Sun distance correction factor ($d^2$). This factor accounts for the seasonal variations in the solar irradiance.
    \item \textbf{Sensor Specifications:} Integration of the band-specific SRFs for both the VNIR and SWIR detectors to ensure accurate spectral convolution.
    \item \textbf{Atmospheric State Variables:} Extraction of Total Column Water Vapour (TCWV), Total Column Ozone ($O_{3}$), and AOD at 550~nm directly from the EnMAP Level-1C (L1C) metadata. These parameters, estimated during the mission's preprocessing stage, are used to constrain the 6S radiative transfer simulations.
\end{itemize}

\section{Software Description and Architecture}
\subsection{Core Engine: Py6S Integration}
To facilitate the interaction between the core Fortran-based code of \textit{6SV} and the processing pipeline, the framework utilises \href{https://py6s.readthedocs.io/en/latest/}{\textit{Py6S}} \cite{wilson2013py6s}. \textit{Py6S} is a high-level Python interface designed to automate the parameterization and execution of 6S simulations. It enables the dynamic definition of complex atmospheric profiles, aerosol models, and acquisition geometries. Specifically, radiative transfer processes are simulated on a per-channel basis, utilising a 2.5~nm spectral sampling interval to characterise the EnMAP sensor's response across its 228 spectral channels.

\subsection{Atmospheric Parameterization via Google Earth Engine Integration}

A key advantage of \textit{6ABOS} is its operational flexibility regarding atmospheric characterisation. While the framework can automatically parse the atmospheric state variables provided in the EnMAP L1C metadata, it also offers the capability to retrieve these parameters from global catalogues via the \textbf{Google Earth Engine (GEE)} API. This dual-source approach represents a significant advantage, allowing users to cross-validate atmospheric inputs or substitute missing metadata with quasi-simultaneous satellite products. The integrated GEE workflow supports the dynamic retrieval of atmospheric parameters from several global datasets:

\begin{itemize}
    \item \textbf{AOD:} Derived from the MODIS Multi-angle Implementation of Atmospheric Correction (MAIAC) product (\texttt{MODIS/061/MCD19A2\_GRANULES}), providing high-resolution aerosol optical depth at 1 km.
    \item \textbf{Total Column Ozone:} Retrieved from the TOMS/SBUV Merged Total Ozone Data (\texttt{TOMS/MERGED}), ensuring a long-term consistent record of atmospheric ozone content.
    \item \textbf{Precipitable Water Vapour:} Obtained from the NCEP/NCAR Reanalysis Surface Water Vapour dataset (\texttt{NCEP\_RE/surface\_wv}), used to characterise the column water vapour content for atmospheric attenuation modelling.
\end{itemize}

Looking forward, future iterations of \textit{6ABOS} are planned to incorporate the retrieval of atmospheric constituents from the \textbf{Copernicus Atmosphere Monitoring Service (CAMS)} reanalysis data, further enhancing the model's reliability in regions with complex aerosol loading or sparse satellite coverage.

\subsection{\textit{6ABOS} Processing Workflow}
\begin{figure}[htbp]
    \centering
    \includegraphics[width=0.9\textwidth]{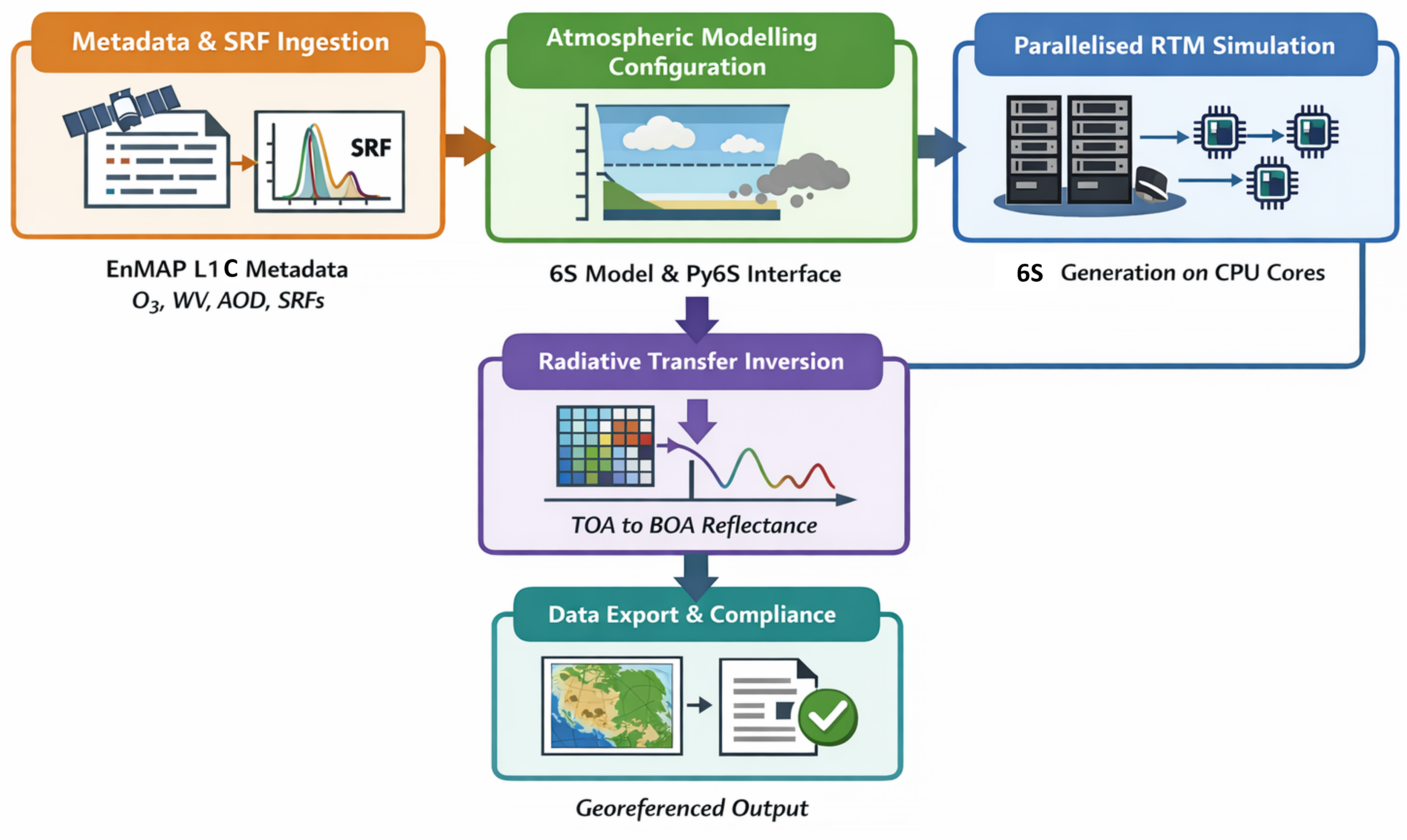}
    \caption{\textit{6ABOS} processing workflow showing the interaction between modules and 6S engine.}
    \label{fig:workflow}
\end{figure}

The \textit{6ABOS} workflow (see Figure~\ref{fig:workflow}) implements a modular five-stage architecture designed to ensure physical consistency and computational efficiency:

\begin{enumerate}
    \item \textbf{Metadata and SRF Ingestion:} Automated parsing of EnMAP L1C metadata to retrieve acquisition geometry, temporal parameters, and atmospheric state variables ($O_{3}$, $WV$, $AOD$). Simultaneously, the sensor-specific SRFs are loaded for spectral convolution.
    \item \textbf{Atmospheric Modelling Configuration:} Dynamic parameterisation of the 6S model using the \textit{Py6S} interface, defining the vertical atmospheric profile and aerosol model.
    \item \textbf{Parallelised RTM Simulation:} The RTM simulations are executed in parallel across multiple CPU cores. This stage computes the atmospheric $L_{path}, T_{g,O_3}, T_{\uparrow}, S_{atm}$, and $E_s$ for all 228 EnMAP spectral channels concurrently.
    \item \textbf{Radiative Transfer Inversion:} Pixel-wise transformation of TOA radiance into BOA reflectance by applying the inverted 6S equation (see Equation~\ref{eq:inversion}).
    \item \textbf{Data Export and Compliance:} Generation of the final georeferenced raster product, ensuring metadata consistency and standard data formats.
\end{enumerate}

\subsection{\textit{6ABOS} Architecture}

The \textit{6ABOS} software package is implemented following a modular “src-layout”, facilitating long-term maintainability and straightforward integration into larger processing frameworks such as the \href{https://github.com/GFZ/enpt}{\textit{EnMAP Processing Tool (EnPT)}}. This design ensures decoupled functionality and promotes the reuse of individual components within broader Earth Observation (EO) workflows.

\subsubsection{Core Components}

The architecture is organised into well-defined modules, each responsible for a distinct aspect of the AC pipeline:

\begin{itemize}
    \item \textbf{\texttt{core.py} (Physics Engine):} Implements the \texttt{SixABOSEngine} class, which encapsulates the RTM logic based on the 6S engine and the core physics underpinning AC. This module orchestrates the preparation of RTM tasks and applies AC to TOA radiance values.

    \item \textbf{\texttt{atmospheric.py} (Atmospheric Data Interface):} Provides an abstraction layer for retrieving dynamic atmospheric state parameters using GEE API based on image acquisition time and geometry. 

    \item \textbf{\texttt{utils.py} (I/O \& Metadata):} Manages ingestion and parsing of EnMAP L1 metadata (XML), handles SRFs retrieval, and supports the generation of standardised georeferenced output products, preserving radiometric and spectral fidelity. 
\end{itemize}

\begin{figure}[htbp]
    \centering
    \includegraphics[width=\textwidth]{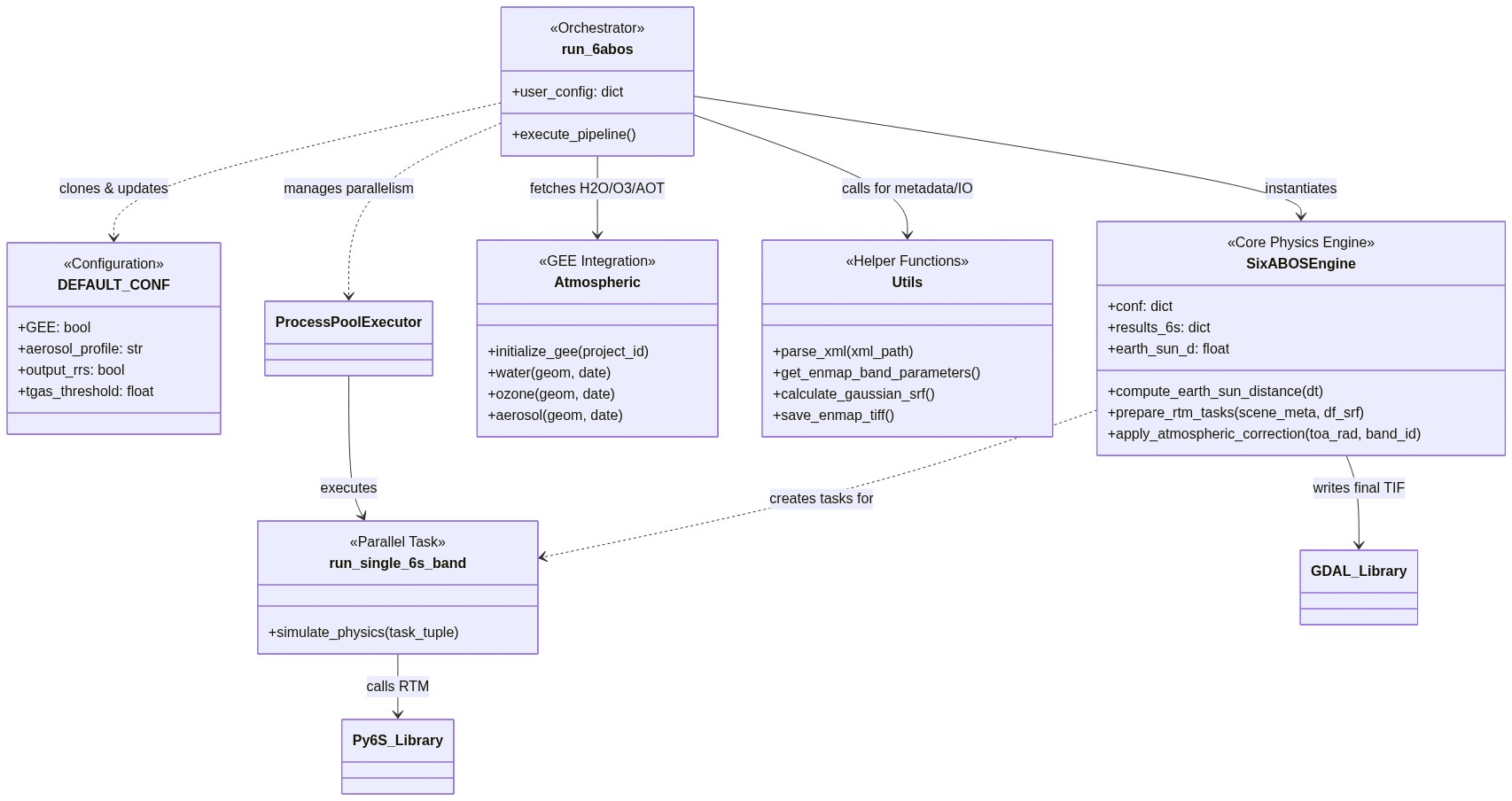} 
    \caption{UML class diagram illustrating the modular interaction between the orchestrator, physics engine, atmospheric interface, and utility modules.}
    \label{fig:uml}
\end{figure}

The software’s modularisation enables clear separation of concerns: the orchestration layer manages configuration parsing and parallel execution (via \texttt{ProcessPoolExecutor}), the physics engine drives the 6S-based correction logic, and auxiliary modules handle data access and external atmospheric data retrieval, facilitating efficient AC workflows tailored to the hyperspectral EnMAP mission. 

\section{Environment Setup and Installation}

To ensure long-term reproducibility and ease of deployment, \textit{6ABOS} is distributed via \textbf{conda-forge}, the community-led repository for the \textit{conda} package manager. This distribution method automatically manages the complex binary dependencies required by the framework, including the 6S radiative transfer core and geospatial libraries.

The environment can be initialized and the framework installed using a single command within a \textit{conda} or \textit{mamba} shell:

\begin{tcolorbox}[colback=gray!5,colframe=gray!50,arc=0pt,outer arc=0pt,top=5pt,bottom=5pt]
\texttt{conda install -c conda-forge sixabos}
\end{tcolorbox}

The source code and the corresponding \textit{feedstock} for the \textit{conda-forge} distribution are openly hosted on GitHub\footnote{\url{https://github.com/conda-forge/sixabos-feedstock}}, allowing for transparent version control and continuous integration of the processing pipeline.

\subsection{Installation Verification}

The successful deployment of the \textit{6ABOS} environment and its integration with the RTM can be verified through the command-line interface. The following log demonstrates the standard procedure for environment creation and the initialisation of the core modules:

\begin{lstlisting}[language=bash, breaklines=true, basicstyle=\small\ttfamily, frame=single, backgroundcolor=\color{gray!10}, keywordstyle=\color{blue}]
# Create and activate the environment
(base):~$ mamba create -n sixabos sixabos
(base):~$ mamba activate sixabos

# Verify module hooking in Python
(sixabos):~$ python
>>> import sixabos
--- [SYSTEM] 6ABOS Modules successfully hooked ---

# Verify CLI accessibility and 6S parameterization options
(sixabos):~$ sixabos-run -h
usage: sixabos-run [-h] [--input INPUT] [--output OUTPUT] 
                   [--aerosol {Continental,Maritime,Urban,Desert,BiomassBurning}]

6ABOS Atmospheric Correction for EnMAP

options:
  -h, --help            show this help message and exit
  --input INPUT         Input folder containing EnMAP data
  --output OUTPUT        Output directory
  --aerosol {Aerosol Models}
\end{lstlisting}

\subsection{Running the 6ABOS Workflow}

The \texttt{6ABOS} AC workflow is executed through the command~\texttt{sixabos-run}. This command processes an EnMAP Level-1C scene and produces atmospherically corrected outputs in a user-defined directory. The processing configuration is controlled through
command-line parameters such as the input dataset, output location, and the selected aerosol model.

To run the workflow, provide the path to the EnMAP Level-1C product using the \texttt{--input} argument and define the destination directory with
\texttt{--output}. The aerosol profile can be specified through the \texttt{--aerosol} parameter. The following example illustrates the execution of the workflow using the Continental aerosol model.

\begin{lstlisting}[
language=bash,
breaklines=true,
breakatwhitespace=false,
columns=fullflexible,
basicstyle=\small\ttfamily,
frame=single,
backgroundcolor=\color{gray!10},
keywordstyle=\color{blue}
]
sixabos-run \
  --input D:\EnMAP_L1\Lake_Garda\Lake_Garda\ENMAP01-____L1C-DT0000083993_20240711T110154Z_008_V010502_20250121T112650Z \
  --output D:\6ABOS_Results\Lake_Garda \
  --aerosol Continental
\end{lstlisting}

\section{Results and Preliminary Validation}

\subsection{Illustrative Performance Example}
\begin{figure}[htbp]
    \centering
    \includegraphics[width=0.9\textwidth]{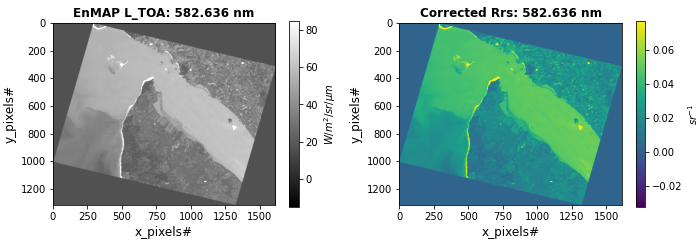}
    \caption{Spatial distribution of the \textit{6ABOS} AC for the EnMAP channel centered at 582.636 nm. Left: $L_{TOA}$ before AC. Right: $R_{rs}$ after AC. The maps present and example of a \textit{6ABOS} output over the Gironde Estuary region, France.}
    \label{fig:ilustrative_example_map}
\end{figure}

The operational implementation of the \textit{6ABOS} framework is demonstrated through its application to an EnMAP L1C scene acquired over the Gironde Estuary (France) on August 1, 2024. This region is characterised by high suspended particulate matter, which provides a challenging yet ideal scenario for validating AC in coastal waters. The spatial consistency of the correction is illustrated in Figure~\ref{fig:ilustrative_example_map}.

\begin{figure}[htbp]
    \centering
    \includegraphics[width=0.85\textwidth]{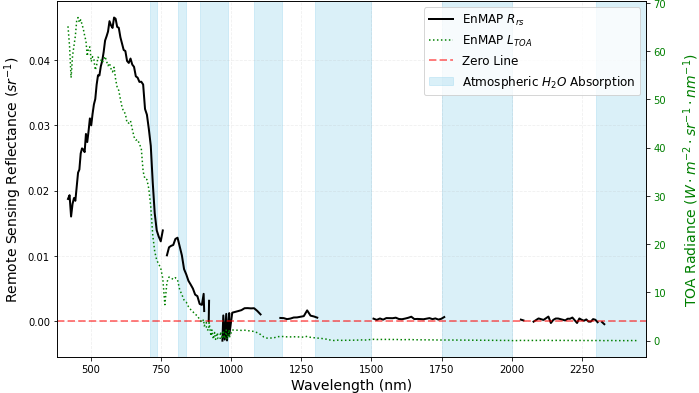}
    
    \vspace{0.4cm} 
    
    \includegraphics[width=0.85\textwidth]{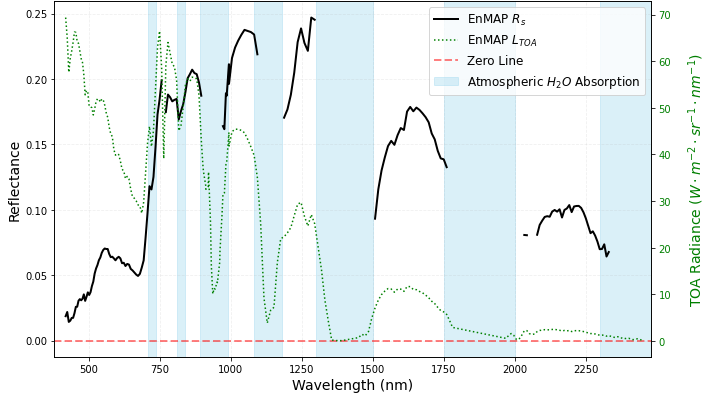}
    
    \caption{Performance of the \textit{6ABOS} AC over the Gironde Estuary (August 1, 2024). Top: Comparison between EnMAP L1C TOA radiance and retrieved $R_{rs}$ after AC for a coastal water pixel. Bottom: surface reflectance of a vegetation pixel in the vicinity of the Gironde estuary. Discontinuities in the spectral curve correspond to regions where total atmospheric transmittance is below the 0.85 threshold. }
    \label{fig:enmap_examples_combined}
\end{figure}

As shown in Figure~\ref{fig:enmap_examples_combined}~(top), \textit{6ABOS} successfully removes path radiance and compensates for the impact of Rayleigh and aerosol scattering, which typically dominate the signal in shorter wavelengths (blue-green region). Regarding atmospheric absorption, the framework employs a mask based on a user-defined total transmittance $\tau_{total}$ threshold (set to 0.85 in this case). Consequently, spectral bands where the total atmospheric transmittance falls below this limit—most notably within the Oxygen-A band ($\sim$760 nm) and major water vapour ($H_2O$) absorption regions—are excluded from the final product output.

Additionally, Figure~\ref{fig:enmap_examples_combined}~(bottom) illustrates the surface reflectance spectrum of a vegetation pixel located near the Gironde Estuary, covering the full VNIR + SWIR spectral range. The $\tau_{total}$, which accounts for both upward and downward gas absorption and scattering processes, is predominantly driven by water vapour absorption in the SWIR region. The resulting spectrum demonstrates the effectiveness of \textit{6ABOS} in recovering the characteristic vegetation "red-edge" and chlorophyll absorption peaks.

\subsection{Case Study: EnMAP Processing over Mediterranean Inland Water Reservoirs}

The performance of~\textit{6ABOS} was evaluated using EnMAP Level 1C imagery acquired over two Mediterranean inland mountain reservoirs in Valencia (Spain) with contrasting optical and environmental properties: Benagéber and Bellús. Benagéber is a deep, oligotrophic mountain reservoir located in the upper reaches of the Turia river basin; its headwater position and minimal upstream human pressure result in high transparency and low chlorophyll-a concentrations, serving as a clear-water reference. Conversely, Bellús is a hypertrophic water body situated in the lower part of the Albaida river basin. Its location makes it subject to intense anthropogenic pressure within its drainage catchment, leading to high nutrient loading, extreme turbidity, and a complex reflectance signal that provides a challenging scenario for AC.

\subsubsection{Validation Datasets}
In situ spectroscopy data were collected at three sampling points within each reservoir using a portable spectroradiometer (ASD FieldSpec® HandHeld 2). The field campaigns were specifically scheduled for April 22 and July 24, 2024, to ensure temporal coincidence with the EnMAP overpasses within a $\pm$3-hour time window. In situ $R_{rs}$ was derived following standard protocols to validate the \textit{6ABOS}-corrected $R_{rs}$ products.

\subsubsection{Validation Results}
\begin{figure}[htbp]
    \centering
    \begin{subfigure}[b]{0.49\textwidth}
        \centering
        \includegraphics[width=\textwidth]{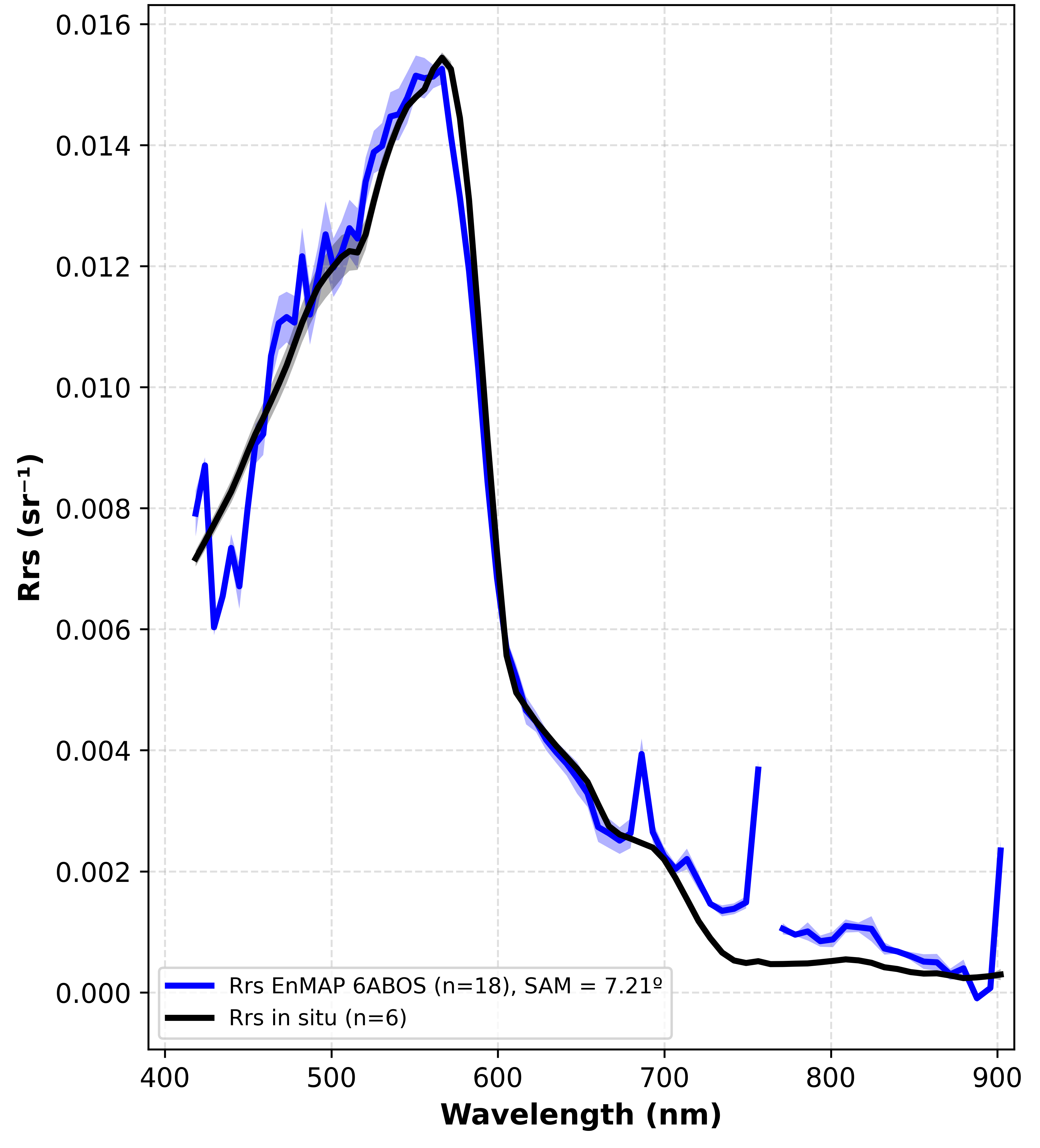}
        \caption{Benagéber oligotrophic reservoir $R_{rs}$.}
        \label{fig:toa}
    \end{subfigure}
    \hfill
    \begin{subfigure}[b]{0.49\textwidth}
        \centering
        \includegraphics[width=\textwidth]{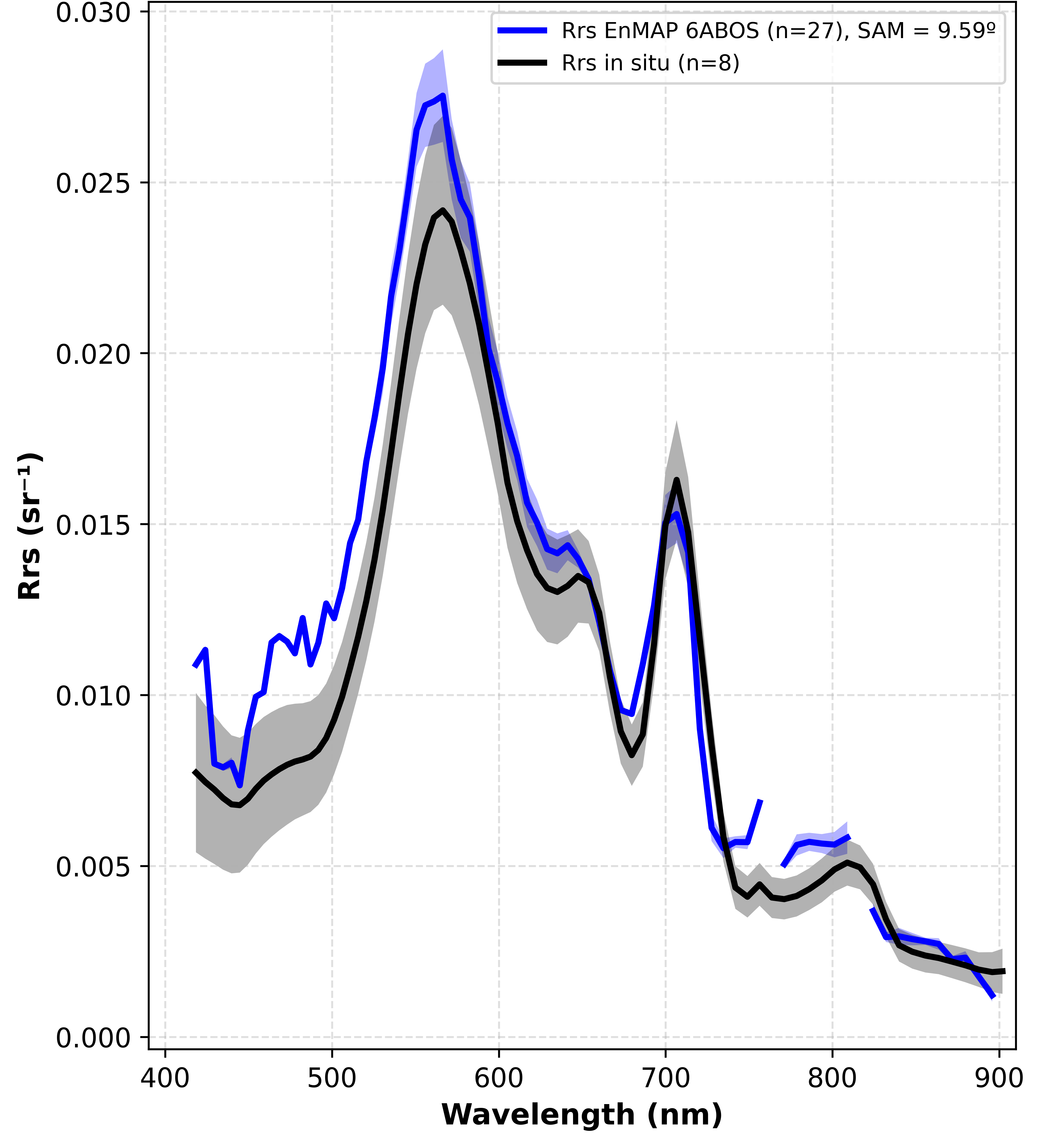}
        \caption{Bellús hypertrophic reservoir $R_{rs}$.}
        \label{fig:boa}
    \end{subfigure}


    \begin{subfigure}[b]{0.49\textwidth}
        \centering
        \includegraphics[width=\textwidth]{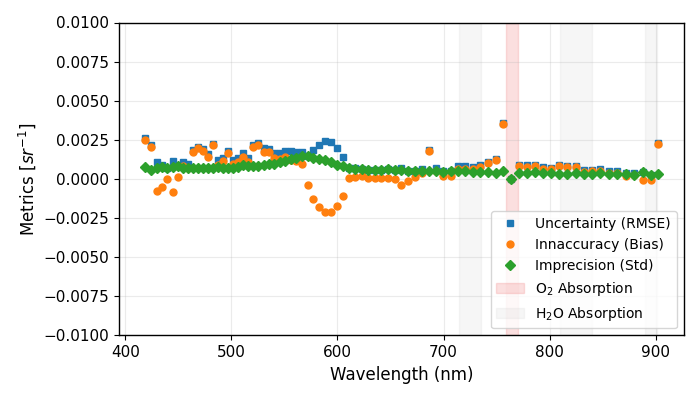}
        \caption{AC lack of performance in Benagéber.}
        \label{fig:fig3}
    \end{subfigure}
    \hfill
    \begin{subfigure}[b]{0.49\textwidth}
        \centering
        \includegraphics[width=\textwidth]{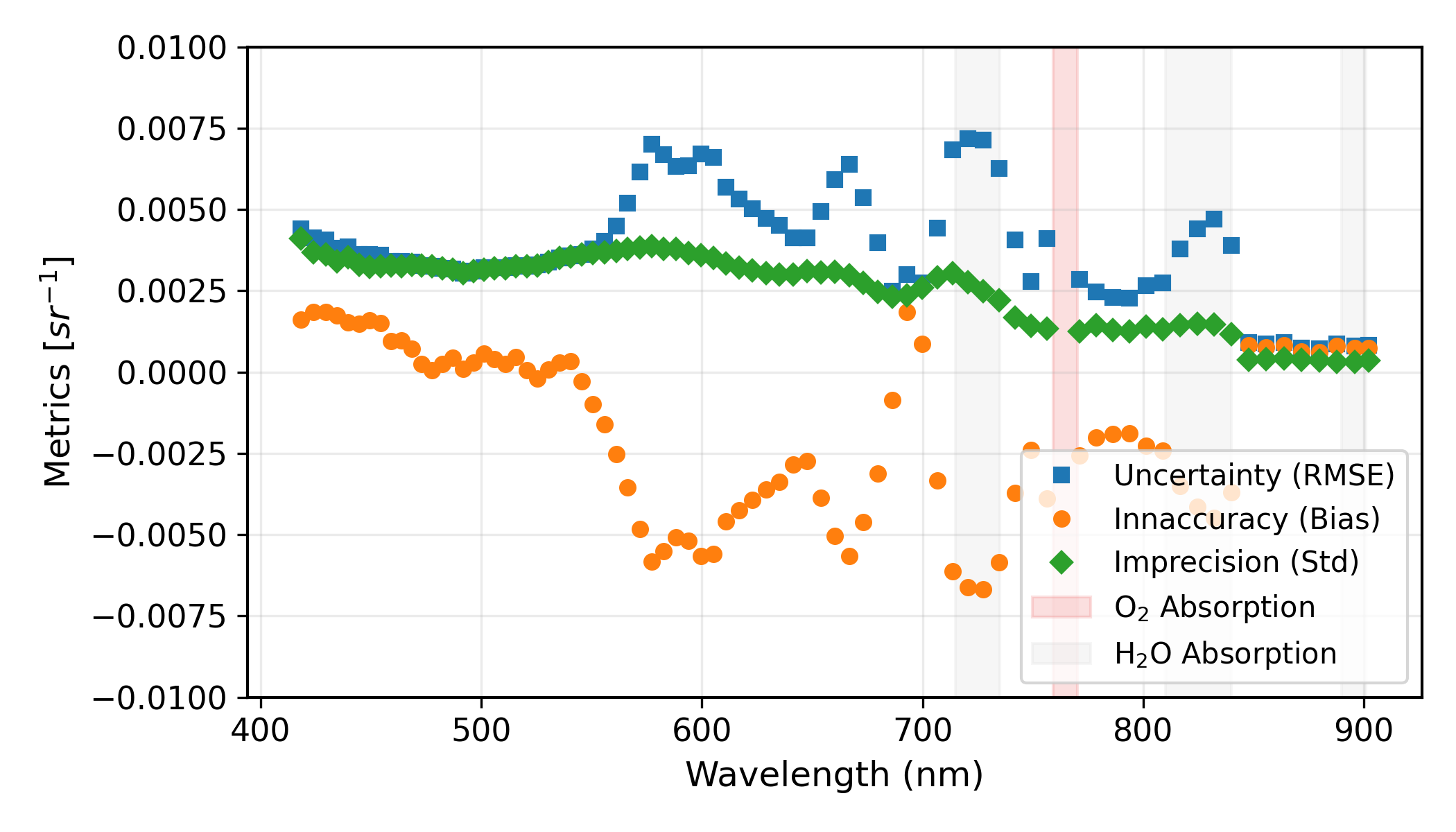}
        \caption{AC lack of performance in Bellús.}
        \label{fig:fig4}
    \end{subfigure}

    \caption{Validation of \textit{6ABOS} AC against in situ measurements. Panels (a) and (b) show the spectral comparison between EnMAP-derived $R_{rs}$ (blue) and in situ $R_{rs}$ (black) for the Benagéber and Bellús reservoirs, respectively. Solid lines represent the mean spectra, while shaded areas indicate the standard deviation ($\pm 1\sigma$). Panels (c) and (d) provide the quantification of the lack of performance for both sites across the 400--900 nm spectral range.}
    \label{fig:results_grid}
\end{figure}

Figure~\ref{fig:results_grid} illustrates the validation of the \textit{6ABOS} framework across both study sites. Panels (a) and (b) compare the in situ $R_{rs}$ (averaged over the three sampling points) with the corresponding $R_{rs}$ derived from the EnMAP L1C data. Although $R_{rs}$ was successfully retrieved for both VNIR and SWIR regions, the results are presented within the 400--900 nm spectral window. This range was selected because water absorbs solar radiation beyond 900 nm; thus, the 400--900 nm interval encompasses the specific regions where the water's Inherent Optical Properties (IOPs) dominate the spectral signature. Panels (c) and (d) provide a quantitative assessment of the AC performance by comparing the \textit{6ABOS} products against the in situ spectroscopy measurements. To quantify the lack of performance, we report the following statistical metrics: uncertainty (Root Mean Square Error, RMSE), inaccuracy (Bias), and imprecision (standard deviation, Std).

The Spectral Angle Mapper (SAM)~\cite{kruse1993spectral} was employed to quantitatively evaluate the similarity between the shape of the EnMAP-derived spectra and the in situ measurements, with lower values indicating higher spectral affinity~\cite{BRAGA202299}. Unlike magnitude-dependent metrics, SAM quantifies spectral similarity by computing the angular difference between two reflectance vectors, effectively measuring the geometry of the spectral profile rather than its magnitude. SAM treats each spectrum as a vector in high-dimensional space and is invariant to scalar changes in amplitude, making it a robust descriptor of spectral shape in hyperspectral remote sensing applications \citep{kruse1993spectral}. This geometric interpretation of spectral similarity is particularly advantageous when comparing $R_{rs}$ spectra, as it captures diagnostic features that are independent of absolute brightness. The SAM results for both study sites, expressed in degrees, are reported in Figures~\ref{fig:results_grid}a and b.

\section{Code Availability}
The \textit{6ABOS} framework is provided as an open-source tool to ensure transparency and reproducibility in hyperspectral data processing. The source code, user manual, and example notebooks are available at the official GitHub repository: \href{https://github.com/PhD-Gabriel-Caballero/6ABOS}{https://github.com/PhD-Gabriel-Caballero/6ABOS}. 

For ease of installation and environment management, the package is also distributed through the \textbf{conda-forge} channel under the name \texttt{sixabos}. The specific version discussed in this paper is archived on Zenodo with the following DOI: \href{https://doi.org/10.5281/zenodo.18232368}{10.5281/zenodo.18232368}.

\section{Impact and Distribution}
\textit{6ABOS} is released under the CC BY 4.0 license to foster open science. By providing a transparent and modifiable codebase, we aim to support the wider hyperspectral community in developing standardised, simple, and robust AC protocols. For ease of use and long-term sustainability, the framework is distributed via GitHub and is available as a \texttt{conda-forge} package, ensuring that these standardised workflows are accessible to researchers across different computing environments.

\section{Conclusions}

This study has presented and validated \textit{6ABOS}, a physically-based AC framework designed for the EnMAP hyperspectral mission. 
By inverting the 6S RTM and integrating automated metadata handling, \textit{6ABOS} significantly reduces the technical complexity and preprocessing time.

The validation conducted over Mediterranean inland waters—ranging from oligotrophic to hypertrophic conditions—demonstrates that the algorithm consistently recovers the spectral signature of complex aquatic environments.

Despite these advancements, AC remains an evolving challenge in remote sensing, particularly regarding the characterisation of extreme aerosol events and adjacency effects in fragmented landscapes. To contribute to these ongoing efforts, \textit{6ABOS} is provided as an open-science tool. Its simple, scalable, and modular architecture is intended to support the scientific community in achieving reproducible and transparent  AC workflows.

\section*{Acknowledgments}

This work was supported by the PROMETEO Program of the Generalitat Valenciana through the RESSBIO project (Remote Sensing Spectroscopy for Wetlands Biodiversity, grant number CIPROM/2021/49). 

The authors also express their gratitude to the EnMAP scientific team and the German Aerospace Center (DLR) for providing the hyperspectral data and technical documentation. 

\bibliographystyle{ieeetr} 
\bibliography{references}

\end{document}